\documentclass[times, twoside]{zHenriquesLab-StyleBioRxiv}
\usepackage{blindtext}
\usepackage{lipsum}
\usepackage[nohyperlinks,printonlyused]{acronym}
\begin{document}

% insert title
\title{A bag of tricks for real-time Mitotic Figure detection}
% insert title for footer
\shorttitle{Approach for MIDOG 2025}

% Use letters for affiliations, numbers to show equal authorship (if applicable) and to indicate the corresponding author
\leadauthor{Christian Marzahl}

\author[1]{Christian Marzahl}
\author[1]{Brian Napora}

\affil[1]{ Gestalt Diagnostics, Spokane WA 99202, USA}

\maketitle

%TC:break Abstract
%the command above serves to have a word count for the abstract
\begin{abstract}
\Ac{mf} detection in histopathology images is challenging due to large variations in slide scanners, staining protocols, tissue types, and the presence of artifacts. This paper presents a collection of training techniques -- a bag of tricks -- that enable robust, real-time \ac{mf} detection across diverse domains. We build on the efficient RTMDet single stage object detector to achieve high inference speed suitable for clinical deployment. Our method addresses scanner variability and tumor heterogeneity via extensive multi-domain training data, balanced sampling, and careful augmentation. Additionally, we employ targeted, hard negative mining on necrotic and debris tissue to reduce false positives. In a grouped 5-fold cross-validation across multiple \ac{mf} datasets, our model achieves an F\textsubscript{1} score between 0.78 and 0.84. On the preliminary test set of the \ac{midog} 2025 challenge, our single-stage RTMDet-S based approach reaches an F\textsubscript{1} of 0.81, outperforming larger models and demonstrating adaptability to new, unfamiliar domains. The proposed solution offers a practical trade-off between accuracy and speed, making it attractive for real-world clinical adoption.

\end{abstract}
%TC:break main
%the command above serves to have a word count for the abstract


\begin{corrauthor}
cmarzahl@gestaltdiagnostics.com
\end{corrauthor}

\section*{Introduction}
\acresetall
Mitotic figure (mf) detection is an important task in evaluating several different cancer types. The use of machine learning based algorithms in digital pathology is emerging as a valuable tool for identifying and quantifying mitotic activity. However, algorithm performance can deteriorate significantly under domain shifts caused by different whole-slide scanners, staining protocols, tissue types, and even species~\cite{midog2021,midog2022}. These variations lead to numerous challenges: e.g., differences in color appearance between laboratories and even scanners, heterogeneous tumor morphology, and distractors such as artifacts, necrotic regions, or inflammation (that can be mistaken for mitoses.) Past challenges like MIDOG 2021~\cite{midog2021} MIDOG 2022~\cite{midog2022} have quantified this problem, with the top-performing algorithm in 2022 achieving an F\textsubscript{1} of 0.76 on unseen domains~\cite{JAHANIFAR2024103132} with a two stage approach. Domain adaptation techniques (such as domain-adversarial training) have been explored to mitigate scanner effects. For example, Wilm \textit{et al.} used a Domain Adversarial RetinaNet to improve generalization in the MIDOG 2021 challenge, reporting an F\textsubscript{1} of 0.71 on the test set~\cite{WilmDomainAdversarial}. These results underscore both the feasibility of domain-generalized \ac{mf} detection and the need for further improvements.

In this work, we propose a high-performance yet real-time solution for \ac{mf} detection that is well-suited for clinical deployment. We build upon the recent RTMDet one-stage detector architecture \cite{lyu_rtmdet_2022}, selecting the small variant (RTMDet-S) for its excellent accuracy-speed trade-off. RTMDet-S has been shown to surpass previous real-time detectors like YOLO in both speed and accuracy \cite{lyu_rtmdet_2022} while maintaining a compact model size. By leveraging this efficient backbone and a suite of training tricks, we aim to achieve robust cross-domain performance without sacrificing inference speed. Our approach includes carefully balanced sampling across multiple datasets, incorporation of mining of hard negative examples from challenging \ac{nmf} regions, and moderate data augmentations to enhance generalization. We demonstrate that these strategies allow a relatively small model to generalize across diverse tumor types and scanners, yielding state-of-the-art results in the latest \ac{mf} MIDOG 2025 detection benchmark. In the following, we detail the utilised publicly available datasets (Section II) and methods (Section III), present evaluation results (Section IV), and discuss the implications of our findings (Section V).

%  Figure \ref{fig:computerNo} shows an example of how to insert a column-wide figure. To insert a  figure wider than one column, please use the \verb|\begin{figure*}...\end{figure*}| environment. Figures wider than one column should be sized to 11.4 cm or 17.8 cm wide. Use \verb|\begin{SCfigure*}...\end{SCfigure*}| for a wide figure with side captions.

\section*{Datasets}

We trained and evaluated our model using a combination of four ~\ac{mf} detection datasets covering a wide range of domains. First, the MIDOG++ dataset \cite{Aubreville2023} provides region-of-interest images from 503 histopathology specimens across seven tumor types (including human breast and lung carcinoma, canine mast cell tumor, melanoma, etc.), with a total of 11,937 \acp{mf}. These images originate from multiple laboratories and scanners, embodying substantial domain variability. We further included two canine tumor datasets: the Canine Cutaneous Mast Cell Tumor (CCMCT)~\cite{Bertram2019} dataset (32 whole-slide images of canine mast cell tumors, fully annotated for mitoses) and a Canine Mammary Carcinoma (CMC)~\cite{Aubreville2020} dataset consisting of 21 whole-slide images of canine breast carcinoma with exhaustive ~\ac{mf} annotations. Both datasets introduce domain shifts in species and tissue type, complementing the human data. Additionally, we incorporated the alternative labeled data from the TUPAC16 challenge (human breast cancer) ~\cite{TUPAC16Alternative} to enrich our training set with more examples of breast tumor morphology. 

To improve the model’s discriminative ability, we performed hard negative mining using two external collections of \ac{nmf} tissue patches. Specifically, we sampled patches from the HistAI SPIDER dataset~\cite{nechaev2025spidercomprehensivemultiorgansupervised} focusing on necrotic labeled tissue regions (necrosis can visually mimic ~\ac{mf}) and on debris labeled patches from the NCT-CRC-HE-100K dataset ~\cite{kather2019predicting} of colorectal tissue \ac{he} images (assuring a variety of structural but \ac{nmf} patterns). These patches, which contain no true \ac{mf}, were added as challenging negative examples during training. By training with this diverse set of domains and hard negatives, we force the model to distinguish \acp{mf} from a broad spectrum of confounders and domain artifacts.

To unify the different annotations and image formats, Gestalt's AI-Studio\footnote{https://www.gestaltdiagnostics.com}(based on EXACT~\cite{marzahl_exact_2021}) was utilized for adjudication and as the central dataset repository.

\section*{Methods}

Our detection model is based on RTMDet~\cite{lyu_rtmdet_2022}, a one-stage anchor-based detector optimized for real-time performance. We adapted it to small object detection by using an appropriate anchor size (64$\times$64 pixels, roughly corresponding to typical ~\ac{mf} size at 40$\times$ magnification) and by leveraging transfer learning from a COCO-pretrained checkpoint. The network was trained on image patches of size 1920$\times$1280 pixels (extracted from whole-slide images at high magnification), which provide enough context around candidate mitoses. Training was done for 25,000 iterations with an effective batch size of 24. Due to GPU memory constraints (using a single NVIDIA RTX 3090), gradient accumulation was utilized to achieve the batch size of 24 over multiple sub-steps for larger RTMDet-based architectures like L, X, ConvNeXt, or SWIN. We used the AdamW optimizer with a warm-up initial learning rate of $1\times10^{-5}$ for 250 steps with a linearly reached peak at $1\times10^{-3}$. Afterward, we adopted a cosine decay learning rate schedule that gradually lowered the learning rate to zero. The detection loss used a Focal Loss (with $\gamma=2$) for the classification branch and an IoU Loss for bounding box regression, which proved effective for the highly class-imbalanced (background candidates, \ac{mf}, \ac{nmf}), small-object detection task. 

To stabilize training and boost final performance, we maintained an \ac{ema} of the model weights. We employed three different \ac{ema} decay rates in parallel ($1\times10^{-3}$, $5\times10^{-4}$, and $25\times10^{-5}$), resulting in three \ac{ema}-weighted models in addition to the primary (non-\ac{ema}) model. 
These four model instances (one original and three EMA variants) were saved at the point of lowest validation loss and subsequently combined as an ensemble for inference. The use of \ac{ema}, especially with multiple decay settings, provided model smoothing which improved generalization and reduced variance between training runs.

\subsection*{Three-Level Hierarchical Sampling Strategy}
An important component of our training procedure is a three-level, balanced sampling strategy designed to limit bias towards any particular domain or class. This approach is inspired by the Generic quadtree sampling strategies proposed by ~\cite{Marzahl2020} to combat class imbalance across \ac{wsi}. At the \textbf{dataset level}(~\eqref{eq:dataset_weight}), each mini-batch is composed by sampling a roughly equal number of training patches from each of the six source datasets (MIDOG++, CCMCT, CMC, TUPAC16, SPIDER, NCT-CRC-HE-100k). This ensures that no single dataset (or tumor type/scanner) dominates the gradient updates. At the \textbf{slide level}(~\eqref{eq:image_weight}), we strive to draw patches evenly from different whole-slide images within each dataset, so that the model sees a broad array of patients and tissue regions. Finally, at the \textbf{patch level}(1920, 1280)(~\eqref{eq:class_weight}), we sample a balance between \ac{mf} and \ac{nmf} instances. In practice, each batch is constructed to have roughly a 1:1 ratio of \ac{mf} vs. \ac{nmf} patches (this includes the hard negative patches from the external sources as part of the \ac{nmf} group). This multi-tiered balancing approach reduces the inherent class imbalance and limits ability of "easy" negative examples to overwhelm the training. It also mitigates domain imbalance by preventing the scenario where, for instance, a batch might consist mostly of images from a single scanner or tumor type. Our ablations determined this strategy was important for achieving consistent performance across all domains, as it forces the model to pay equal attention to mitoses in every context.

Let $x_i$ denote the $i$-th sample in the dataset of a total of 112.923 patches (1920, 1280). The combined sampling weight $w(x_i)$ ~\eqref{eq:combined_weight} is defined as the product of three distinct weights: dataset-level, image-level, and class-level.

\begin{equation}
    w(x_i) = w_{D}(d_i) \cdot w_{I}(m_i) \cdot w_{C}(c_i)
    \label{eq:combined_weight}
\end{equation}

where:

\begin{itemize}
    \item $d_i$ is the dataset containing sample $x_i$.
    \item $m_i$ is the slide from dataset $d_i$ containing sample $x_i$.
    \item $c_i$ is the class label (\ac{mf} or \ac{nmf}) of sample $x_i$.
\end{itemize}

The individual weights are defined as follows:

\textbf{Dataset-level weight} ($w_D$):
\begin{equation}
    w_{D}(d) = \frac{\frac{1}{|d|}}{\sum_{j=1}^{|D|}\frac{1}{|d_j|}}
     \label{eq:dataset_weight}
\end{equation}
where $|d|$ is the number of samples in dataset $d$, and $D$ is the set of all datasets.

\textbf{Image-level weight} ($w_I$):
\begin{equation}
    w_{I}(m) = \frac{\frac{1}{|m|}}{\sum_{k=1}^{|M|}\frac{1}{|m_k|}}
    \label{eq:image_weight}
\end{equation}
where $|m|$ is the number of annotations in image $m$, and $M$ is the set of all images across all datasets.

\textbf{Class-level weight} ($w_C$):
\begin{equation}
    w_{C}(c) = \frac{1}{N_c}
    \label{eq:class_weight}
\end{equation}
where $N_c$ is the total number of samples belonging to class $c$ (\ac{mf} or \ac{nmf}).

This hierarchical weighting strategy ensures balanced representation across datasets, images, and classes, preventing biases from datasets of varying sizes, annotation-rich images, or imbalanced class distributions.

\subsection*{Augmentation}

We applied data augmentations to expose the model to minor variations in appearance and orientation, improving its ability to handle differences in staining and imaging. Each training patch underwent random horizontal and vertical flips with 50\% probability, as well as a small random rotation uniformly sampled from -15$^\circ$ to +15$^\circ$. These geometric augmentations account for the fact that \acp{mf} can appear in any orientation. In addition, we performed color jittering in the HSV color space: we randomly adjusted the hue, saturation, and value of the image within a limited range (e.g. $\pm10\%$ hue variation, $\pm20\%$ saturation, $\pm20\%$ brightness). This simulates variations in staining (H\&E dye concentration, scanner white balance, etc.) without altering the underlying tissue morphology. We avoided excessive or unrealistic augmentations (such as heavy blurring or artificial noise) since preliminary experiments showed that preserving histological details was important for \ac{mf} detection. The chosen augmentations yielded a modest boost in validation performance and helped the model maintain stable predictions across slides from different sources. We abandoned the use of MixUp and CutMix after introducing hard negative mining, in order to avoid generating training examples that place \ac{mf} on necrotic or debris tissue.

\subsection*{Hard Negative Mining on Necrotic or Debris Tissue}

To further improve model robustness against false positives, we performed targeted hard negative mining on public datasets containing necrotic or debris-labeled regions. Specifically, we applied our trained RTMDet-S model to the HistAI SPIDER dataset and the NCT-CRC-HE-100K colorectal dataset. During this process, we set a deliberately low detection confidence threshold of 0.25 to capture a broad set of candidate detections, including many spurious objects. All predictions located in necrotic regions or tissue debris, which by definition cannot contain true \ac{mf}, were labeled as \ac{nmf} examples and added to our training pool as hard negatives. This strategy exposed the model to a wide range of mitosis-like imposters (e.g., fragmented nuclei, condensed chromatin remnants, or irregular debris structures) and taught it to suppress confident predictions in such regions. Incorporating these mined negatives proved particularly effective in reducing false positives on challenging test slides containing necrosis, thereby enhancing overall generalization across domains.

\section*{Evaluation and Results}
We evaluated our approach using F\textsubscript{1}-score, the harmonic mean of precision and recall, as the primary metric. A grouped, stratified 5-fold cross-validation was conducted on the combined training dataset, where slides were partitioned such that each fold had a representative mix of all datasets and tumor types. This cross-val evaluation reflects the model’s ability to generalize to unseen slides across all domains. Using the ensemble of four \ac{ema} models and applying test-time augmentation (horizontal flip and 90$^\circ$ rotations at inference), we obtained per-fold F\textsubscript{1} scores ranging from 0.78 up to 0.84. The variance in fold performance was attributable mainly to certain domains being more challenging (e.g., folds where many rare tumor type images were in the test split tended to have the lower scores). Nevertheless, all folds demonstrated F\textsubscript{1} above 0.78, indicating consistently high detection quality.

For the MIDOG 2025 challenge evaluation (Track 1: MF detection across unseen domains), we submitted the predictions of our \ac{ema} ensemble with test-time augmentation. On the preliminary leaderboard, our approach achieved an F\textsubscript{1} score of 0.81, recall 0.84, precision 0.78, placing among the top entries. This result underscores the strong domain generalization of our model, as the challenge test set potentially included new scanners and tumor types not present in training. Importantly, our method accomplishes this performance with real-time inference speed: on an NVIDIA 3090 GPU, the RTMDet-S model processes a 1920$\times$1280 image patch in $\sim$20--30 ms (equating to over 30 patches per second, easily scalable to whole-slide processing within a few minutes).

Interestingly, we found that larger versions of the detector did not provide meaningful improvement. We experimented with the RTMDet-(M,L,X) architecture (larger-capacity models) and an RTMDet variant using a Swin-L Transformer or the ConvNeXt backbone, expecting higher accuracy. However, both alternatives yielded F\textsubscript{1} scores on the cross-validation folds that were on par with or slightly worse than the RTMDet-S. The increased model complexity appeared to overfit the training domains despite regularization, and added computational cost without clear benefit. Thus, the small model struck the best balance between bias and variance for this task. We attribute this to the fact that these architectures were developed to handle the 80 MSCOCO classes and not a binary \ac{mf} vs \ac{nmf} use case. 

We also tested several additional training techniques that, perhaps surprisingly, turned out to be ineffective for our scenario. First, we tried mosaic augmentation (combining multiple image patches into a single training image, as popularized in YOLO frameworks) to provide more context and diversify training samples. This did not improve performance; on the contrary, it sometimes confused the model with unnatural juxtapositions of tissue. Second, we applied stain normalization using standard methods (e.g., Macenko normalization) to harmonize color differences between domains. This had negligible impact on the results, possibly because our HSV jitter already covered color variation or because the model learned stain-invariant features given enough raw data. Lastly, we implemented a detached two-stage detector pipeline: the RTMDet first-stage outputs were used to crop small candidate regions, which were then classified by a separate CNN classifier to verify true \acp{mf}. We tried various backbone networks for this second stage (including EfficientNet and Vision Transformer models), aiming to filter false positives. However, this approach did not boost the final F\textsubscript{1} score and in some cases reduced it. The single-stage detector alone proved sufficient when properly trained, and the added complexity of a second stage was not justified.

\section*{Discussion and Conclusion}

Our study demonstrates that a compact one-stage detector, when paired with appropriate training tricks, can achieve excellent generalization in \ac{mf} detection across a wide range of domains. By combining diverse datasets and addressing class and domain imbalance through equitable sampling, our RTMDet-based model detected variations in tumor type, species, and imaging conditions. The resulting F\textsubscript{1} score of 0.81 on the MIDOG 2025 preliminary test set is, to our knowledge, one of the highest reported for this challenging task. 

In addition to accuracy, a key advantage of our approach is its real-time inference capability. Fast detection is crucial for clinical deployment, where whole-slide images are gigapixel-sized and numerous patches must be processed. Our choice of a small model and efficient one-stage design was validated by the fact that bigger models did not yield gains, reinforcing the idea that execution speed can be achieved without sacrificing accuracy in this domain. This opens the door for integrating automated \ac{mf} detection into the pathologist’s workflow, providing quick second opinions or triaging slides based on proliferative activity.

Our negative findings (mosaic augmentation, stain normalization, second-stage classification) are also informative. They suggest that, for \ac{mf} detection under domain shift, carefully curating the training data distribution and using a well-tuned one-stage detector may be more effective than adding complexity or pre-processing. The failure of the second-stage classifier in particular warrants further analysis.  One explanation could be that the classifier struggled with a lack of tissue context so may not have adequately incorporated surrounding necrotic tissue or general tissue debris.

In conclusion, we presented a practical and high-performing solution for real-time \acp{mf} detection. Using a "bag of training tricks", our method achieved robust cross-domain generalization and top-tier accuracy on a challenging multi-domain benchmark, all while running efficiently. Ongoing enhancements include extending this approach to staining artifacts and pen marking and also examining the applicability to other histopathology tasks. We believe that our findings and approach will help pave the way for reliable, fast AI assistants in pathology labs, ultimately improving the objectivity and reproducibility of mitosis quantification and tumor grading.

%\begin{figure}%[tbhp]
%\centering
%\includegraphics[width=\linewidth]{Figures/mitosis}
%\caption{Placeholder image.}
%\label{fig:computerNo}
%\end{figure}

\begin{acknowledgements}
The authors thank the computational pathology community for providing publicly available datasets.
\end{acknowledgements}

\section*{Bibliography}
\bibliography{literature}

\begin{acronym}
  \acro{ema}[EMA]{Exponential Moving Average}
  \acro{wsi}[WSI]{whole-slide image}
  \acro{mc}[MC]{mitotic count}
  \acro{mf}[MF]{mitotic figure}
  \acro{nmf}[NMF]{non-mitotic figure}
  \acro{fov}[FOV]{field of view}
  \acro{roi}[ROI]{region of interest}
  \acro{he}[H\&E]{hematoxylin and eosin}
  \acro{ap}[AP]{Average Precision}
  \acro{map}[mAP]{mean Average Precision}
  \acro{iou}[IoU]{intersection over union}
  \acro{midog}[MIDOG]{MItosis DOmain Generalization}
  \acro{cnn}[CNN]{convolutional neural network}
\end{acronym}

\end{document}